# Face Recognition Algorithms Based on Transformed Shape features

Sambhunath Biswas[1] and Amrita Biswas[2]

[1] Machine Intelligence Unit, ISI
Kolkata-700108, India

[2] Electronics & Communication Department, Sikkim Manipal Institute of Technology
Majitar, 737132, India

**Abstract**
Human face recognition is, indeed, a challenging task, especially under the illumination and pose variations. We examine in the present paper effectiveness of two simple algorithms using coiflet packet and Radon transforms to recognize human faces from some databases of still gray level images, under the environment of illumination and pose variations. Both the algorithms convert 2-D gray level training face images into their respective depth maps or physical shape which are subsequently transformed by Coiflet packet and Radon transforms to compute energy for feature extraction. Experiments show that such transformed shape features are robust to illumination and pose variations. With the features extracted, training classes are optimally separated through linear discriminant analysis (LDA), while classification for test face images is made through a k-NN classifier, based on $L1$ norm and Mahalanobis distance measures. Proposed algorithms are then tested on face images that differ in illumination, expression or pose separately, obtained from three databases, namely, ORL, Yale and Essex-Grimace databases. Results, so obtained, are compared with two different existing algorithms. Performance using Daubechies wavelets is also examined. It is seen that the proposed Coiflet packet and Radon transform based algorithms have significant performance, especially under different illumination conditions and pose variation. Comparison shows the proposed algorithms are superior.
*Keywords:* Face Recognition, Radon Transform, Wavelet Transform, Linear Discriminant Analysis.

## 1. Introduction

Face Recognition problem has been studied extensively for more than twenty years but even now the problem is not fully solved. In particular, the problem still exists when illumination and pose vary significantly. Recently, some progress [1] has been made on the problems of face recognition, especially under conditions such as small variations in lighting and facial expressions or pose. Of the many algorithms for face recognition, so far developed, the traditional approaches are based on Principal Component Analysis (PCA). Hyeonjoon Moon et al. [2] implemented a generic modular PCA algorithm where the numerous design decisions have been stated explicitly. They experimented with changing the illumination normalization procedure and studied its effect through the performance of compressing images with JPEG and wavelet compression algorithms. For this, they varied the number of eigen vectors in the representation of face images and changed the similarity measure in the classification process. Kamran Etemad and Rama Chellappa in their discriminant analysis algorithm [3], made an objective evaluation of the significance of visual information in different parts (features) of a face for identifying the human subject. LDA of faces provides a small set of features that carries the most relevant information for classification purposes. The features are obtained through eigen vector analysis of scatter matrices with the objective of maximizing between-class variations and minimizing within-class variations. The algorithm uses a projection based feature extraction procedure and an automatic classification scheme for face recognition. A slightly different method, called the evolutionary pursuit method, for face recognition was de-scribed by Chengjun Liu and Harry Wechsler [4]. Their method processes images in a lower dimensional whitened PCA subspace. Directed but random rotations of the basis vectors in this subspace are searched by Genetic Algorithm, where evolution is driven by a fitness function defined in terms of performance accuracy and class separation. Up to now, many face representation approaches have been introduced including subspace based holistic features and local appearance features [17]. Typical holistic features include the well-known principal component analysis (PCA)[18], linear discriminant analysis[19], independent component analysis (ICA)[20] etc. Recently, information from different sections, such as, scale, space and orientation, has been used for representation and recognition of human faces by Zhen et al. [21]. This does not include the effect of illumination change. Subspace based face recognition under the scenarios of misalignments and/or image occlusions has been published by Shuicheng et al. [22]. We have not considered image occlusions as our objective is different. The proposed research work addresses the problem of face recognition to achieve high performance in the face recognition system. Face Recognition method, [5] based on curvelet based PCA and tested on ORL Database, uses 5 images for training and has achieved 96.6% recognition rate and, using 8 images for training on the Essex Grimace database, has achieved 100% recognition rate. Another algorithm [6], based on wavelet transform, uses 5 images for training from the ORL Database has achieved a recognition rate of 99.5%. But still more improvement is required to ensure that the face recognition algorithms are robust, in particular to illumination and pose variation. A face recognition algorithm, mainly based on two dimensional graylevel images, in general, exhibits poor performance when exposed to different lighting conditions. This is because the features extracted for classification are not illumination invariant. To get rid of the illumination problem, we have used the 3-dimensional depth images of the corresponding 2-dimensional graylevel face images. This is because the 3-D depth image depicts the physical surface of the face and thus, provides the shape of human face. The primary reason is that such a shape depends on the gradient values of the physical surface of the face, i.e., on the difference of intensity values and not on the absolute values of intensity. As a result, change in illumination does not affect the feature set and so the decision also remains unaffected. Such a shape can be obtained using a shape from shading algorithm and subsequently can be used for feature extraction. 3-D face matching using isogeodesic stripes through a graph as described in [23] is a different technique for face recognition. But it is computationally expensive.However, it is also a different area of research. Xiaoyang and Triggs [24], on the other hand, considered texture features for face recognition under difficult lighting conditions. Their method needs to enhance local textures but how to select the local textures or which local textures are adequate and need be considered are not discussed. The proposed algorithms use the shape from shading algo-rithm [8], and Coiflet/wavelet and Radon transform respectively to compute energy for feature extraction. It should be noted that Radon transform provides directional information in image. Since, PCA eigen axis provides the direction of maximum variance in data, this axis must rotate with the rotation of the image to maintain the optimality in data variance and hence the DFT magnitude of the directional information, provided by the Radon transform computed with respect to PCA eigen axis gives robust features with respect to rotation. Note that, linear discriminant analysis (LDA) groups the similar classes in an optimal way in the eigen space and so the k-NN classifier can be used for classification. We have used $L1$ norm and Mahalanobis distance to test for classification. With this, the outline of the paper is described as follows: In section 2, we briefly review a shape from shading algorithm and in section 3, the concepts of wavelet packet decomposition, Radon transform and LDA methodology are briefly sketched. Section 4, depicts the proposed two different algorithms, while experimental results and comparison of both the methods are discussed in section 5. Finally, conclusion is made in section 6.

**2. Extraction of Illumination Dependent Features**

The problem of recovering 3-D shape from a single monocular 2-D shaded image was first addressed by B. K. P Horn [14]. He developed a method connecting the surface gradient (p, q) with the brightness values for Lambertian objects. Theresult is known as the reflectance map. Therefore, he computed the surface gradients (p, q) using the reflectance map in order to get the shape. From (p, q), he also computed depth, Z.Since, orientation of tangent planes is accompanied by the orientation of their normal vectors, say $(n_x, n_y, n_z)$, they can also be effectively used to represent the surface shape.As the reflectance map, in general, is non-linear, it is very difficult to find the gradient values in a straightforward way.Some other researchers, such as, Bruss [15] and Pentland [16], to simplify the problem, thought of local analysis to compute the shape. Thus, two different kinds of algorithms,e.g. global and local emerged. In global methods, Horn showed the shape can be recovered by minimizing some cost function involving constraints such as smoothness. He used variational calculus approach to compute the shape in the continuous domain and its iterative discrete version in the discrete domain.Bruss showed that no shape from shading technique can provide a unique solution without additional constraint. Later on, P. S. Tsai and M. Shah [8] provided a simple method to compute shape through linearization of Horn's nonlinear reflectance map. For our purpose, we have used the shape from shading algorithm described by P. S. Tsai and M. Shah [8] for its simplicity and fastness.This approach employs discrete approximations for p and q using finite differences, and linearizes the reflectance in Z(x,y). The method is fast,since each operation is purely local. In addition, it gives good results for the spherical surfaces, unlike other linear methods.Note that the illumination change may be due to the position change of the source keeping the strength of the source as it is or due to the change in the source strength keeping the position of the source fixed. In either case, the gradient values, $p$ and $q$, of the surface do not change, i.e., they can be uniquely determined [14]. Hence, for the linear reflectance map, the illumination will have no effect on the depth map. In other words, depth map will be illumination invariant.

**3. Feature Extraction**

A number of methods are available to extract the facial features. In the proposed methods, we have selected two approaches. One is based on discrete Coiflet packet transform and the other is based on Radon transform. Both these trans-forms use the depth values of face images to extract features for classification.

3.1 Discrete Wavelet Transform

Discrete wavelet transform is a powerful technique in signal processing and can be used in different research areas. Wavelet transform has merits of multi-resolution, multi-scale decomposition, and so on. In frequency field, when the facial image is decomposed using two dimensional wavelet transform, we get four regions. These regions are: one low-frequency region LL (approximate component), and three high-frequency regions,namely LH (horizontal component), HL (vertical component),and HH (diagonal component), respectively. In wavelet packet decomposition [13], we divide each of these four regions in the similar way. The sketch map of wavelet packet decomposition is shown in Fig. 1. As illustrated in Fig. 1, the L denotes low frequency and the H denotes high frequency. With each level of decomposition, the number of pixels in each subband greatly reduces into packets while the essential features of the underlying image are retained. The low frequency region in decompositions at different levels is the blurred version of the input image, while the high frequency regions contain the finer detail or edge information contained in the input image. To ensure almost nearly rotational invariance,the linear combination of the four subbands can be taken.This combination provides the sum of the different energy bands. For coiflets, this linear combination of four subband coiflet coefficients provides excellent constancy when the subject undergoes rotation. We have examined both coiflet and Daubechies wavelet in the proposed first algorithm. Also,note that coiflet has zero wavelet and zero scaling function moments in addition to one zero moment for orthogonality.As a result, a combination of zero wavelet and zero scaling function moments used with the samples of the signal may give superior results compared to wavelets with only zero wavelet moments [7]. This fact also is reflected in our results.

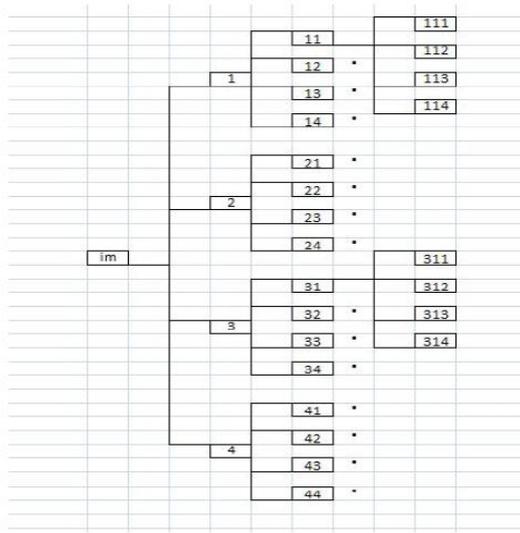

Fig.1 Wavelet Packet Decomposition

### 3.2 Radon Transform and robust features

In the proposed second algorithm, Radon transform is used to derive the linear features. Due to inherent properties of Radon transform, it is a useful tool to capture the directional features of images. It should be noted that the principal axis of PCA for an image rotates when the image rotates. This is done to maintain the maximum variance in data. Therefore,Radon transform, computed with respect to this axis, tenders robust features. The Radon transform of a two dimensional function f(x,y) is defined as

$$R(r,\theta)[f(x,y)] = \int_{-\infty}^{\infty}\int_{-\infty}^{\infty} f(x,y)\delta(r - x\cos\theta - y\sin\theta)dxdy \quad (1)$$

where $\delta(.)$ is the Dirac function, $r \in [-\infty, \infty]$ is the perpendicular distance of a line from the origin and $\theta \in [0, \pi]$ is the angle formed by the distance vector as shown in fig. 2.Radon transform is defined for an image with unlimited support. In practice, the image is confined to [-L, L]x[-L, L]. According to Fourier slice theorem, this transformation is invertible. Fourier slice theorem states that for a 2-D function, the 1-D Fourier transforms of the Radon transform along r, are the radial samples of the 2-D Fourier transform of f(x,y) at the corresponding angles. We know the Radon transform changes as the image rotates. Rotation of the input image corresponds to the translation of the Radon transform along ([9], [11]). Jafari-Khouzani and Soltanian-Zadeh [10] showed how the rotation of the texture sample corresponds to a circular shift along θ. Therefore, using a translation invariant wavelet transform along θ, they produced rotation invariant features. Their method is very expensive in the sense that they used the Radon transform for 1-degree to 180-degree to find the principal direction of texture. In the proposed algorithm, we have used Radon transform along the principal eigen axis given by the PCA method to compute the projections of all training images. Thus, the principal direction here is the direction of PCA eigen axis. The features so obtained are rotationally robust because the principal eigen axis, given by the PCA method, considers always the line that best fits the data cloud. Then their DFT magnitude may be taken to constitute the feature vectors. Thus, directional facial characteristics are incorporated in the feature values.

### 3.3 Linear Discriminant Analysis

Linear Discriminant Analysis (LDA) is a technique commonly used for data classification and dimensionality reduction. LDA easily handles the case where within class frequencies are unequal. It maximizes the ratio of between class variance to within class variance in any particular data set thereby guaranteeing maximal separability. However in face recognition problem, one is confronted with the difficulty that the within class scatter matrix is always singular. This stems from the fact that the number of images in the training set is much smaller than the number of pixels in the image. In order to overcome this problem the images are projected to a lower dimensional space so that the resulting within class

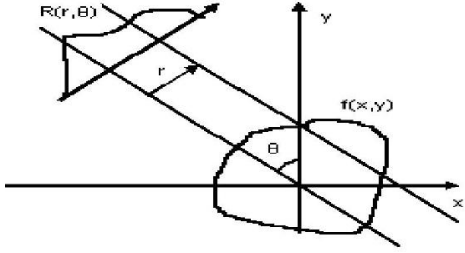

Fig.2 Radon Transform

scatter matrix is non singular. Let the between class scatter matrix be defined as

$$S_B = \sum_{i=1}^{c} N_i (\mu_i - \mu)(\mu_i - \mu)^T, \quad (2)$$

and the within class scatter matrix be defined as

$$S_W = \sum_{i=1}^{c} \sum_{x_k \in X_i} (x_k - \mu_i)(x_k - \mu_i)^T. \quad (3)$$

where $\mu_i$ is the mean image of the class $X_i$ and $N_i$ is the no.of samples in the class Xi. If $S_W$ is non singular the optimal projection $W_{opt}$ is chosen as the matrix with orthonormal columns which maximizes the ratio of the determinant of the between class scatter matrix of the projected samples to the determinant of the within class scatter matrix of the projected samples, i.e,

$$\begin{aligned} W_{opt} &= arg\ max_W \frac{|W^T S_B W|}{|W^T S_W W|}, \\ &= [w_1 w_2 .... w_m]. \end{aligned} \quad (4)$$

where $\{w_i | i=1,2,…m]$ is the set of generalized eigen vectors of $S_B$ and $S_W$ corresponding to the m largest generalized eigen values $\lambda_i |i=1,2,..m$ i.e.,
$S_B w_i = \lambda_i S_W w_i$, i = 1, 2…..m.
Note that there are at most c-1 non zero generalized eigen values and so, an upper bound on m is c-1 where c is the number of classes. In order to overcome the problem for singularity of $S_W$, the regularization technique has been applied. A constant μ > 0 is added to the diagonal elements of $S_W$ as $S_W + \mu I_d$, where $I_d$ is the identity matrix of size 'd'. It is easy to verify that $S_W + \mu I_d$ is positive definite and hence is non singular.

## 4. Proposed Approach

We discuss in this section the proposed two different algorithms, using discrete wavelet transform and Radon transform. The first algorithm uses the discrete Coiflet/Daubechies wavelet for feature extraction and described as follows:

Algorithm 1

Step 1: compute depths of all the training images using shape from shading algorithm.
Step 2: do the four levels of decomposition of depth images using wavelet transform, based on coiflet mother wavelet to get four subband components.
Step 3: take the linear summation of all the wavelet transform coefficients to build up the feature vectors for training images.
Step 4: do the linear discriminant analysis on the feature vectors.
Step 5: compute the feature vectors for test images and project it in the LDA subspace.
Step 6: classify the test images using $L1$ norm and Mahalanobis distance measure.
Step 7: stop.
In Algorithm 2, Radon transform is followed by Fourier transform to capture the directional features of the depth map images. The selected angle for Radon transform computation is θ, detected by the principal eigen axis because of its uniqueness.

Algorithm 2

Step 1: compute depths of all the training images using shape from shading method.
Step 2: compute the Radon Transform coefficients of the depth images and take them as feature vectors or find the DFT magnitude of the transformed data to constitute the feature vectors.
Step 3: do the linear discriminant analysis on the feature vector set.
Step 4: compute the feature vectors for test images and project it in the LDA subspace.
Step 5: classify the test images using the $L1$ norm and Mahalanobis distance measure.
Step 6: stop.

## 5. Results & Discussion

In order to test the proposed algorithms, we have used ORL, Essex-Grimace and Yale databases. ORL (AT and T) database contains 10 different images (92 x 112), each of 40 different subjects. All images were taken against a dark homogenous background with the subjects in upright, frontal position with some side movement. Sample images of the dataset are shown in Fig. 5. Yale Database contains images of 10 subjects(480x 640) under 64 different lighting conditions. Sample images of this database are shown in Fig. 6. Essex-Grimace database, on the other hand, contains a sequence of 20 images (180x200), each of 18 individuals consisting of male and female subjects, taken with a fixed camera. During the sequence, the subject moves his/her head and makes grimaces which get more extreme towards the end of the sequence. Sample images of this database are shown in Fig. 7. The depth map was computed for all the images in the training database assuming the reflectance of the surface to be Lambertian.

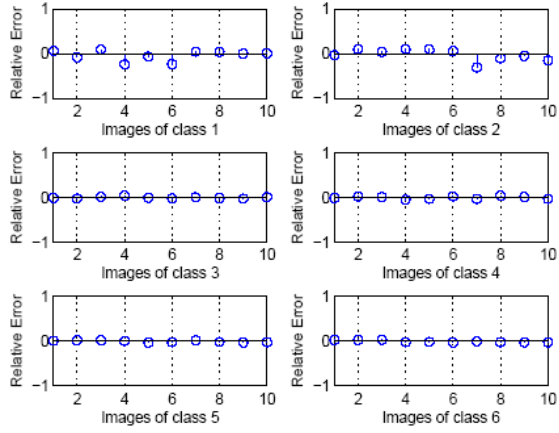

Fig.3 Relative error of images from the respective class mean in Algorithm 1

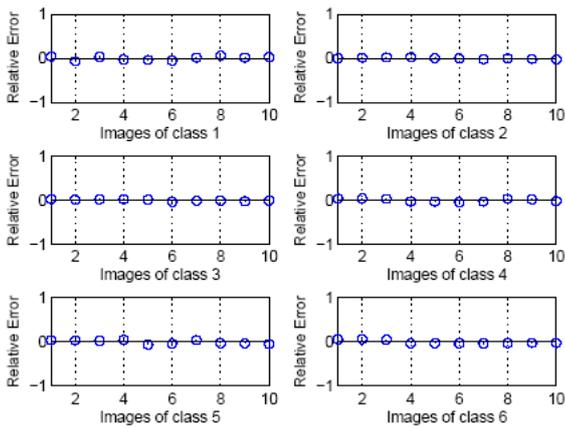

Fig. 4 Relative error of images from the respective class mean in Algorithm2.
.

Table 1:Coiflet Based Classification(Algorithm 1)

| Sl.No. | Data-base | No.of Tr.Images | Classi-fication | Recog-nition % |
|---|---|---|---|---|
| 1 | ORL | 4 | L1 norm | 100.0 |
| 2 | ORL | 4 | Mahalanobis | 100.0 |
| 3 | ORL | 3 | L1 norm | 98.5 |
| 4 | ORL | 3 | Mahalanobis | 98.5 |
| 5 | ORL | 2 | L1 norm | 97.5 |
| 6 | ORL | 2 | Mahalanobis | 95.0 |
| 7 | Yale | 2 | L1 norm | 100.0 |
| 8 | Yale | 2 | Mahalanobis | 100.0 |
| 9 | Essex-Grimace | 2 | L1 norm | 100.0 |
| 10 | Essex-Grimace | 2 | Mahalanobis | 100.0 |
| 11 | Essex-Grimace | 1 | L1 norm | 100.0 |
| 12 | Essex-Grimace | 1 | Mahalanobis | 79.0 |

The obtained depth image has the same size as the original image i.e. 92 x112, 180 x 200 and 480 x 640 for the ORL, Essex-Grimace and Yale Database, respectively. Depth image computed by shape from shading algorithm for the first image in ORL database is shown in Fig. 8.

To show the robustness of features against orientation, we have plotted, in both the cases (Algorithm 1 and Algorithm 2), the relative error in distance measurement for all ten images in six classes (of ORL database) from their respective mean images in the LDA space. Note that this distance is almost zero for all the images in a class and maintains excellent constancy. Also for robustness of features against illumination, we have computed this distance using Yale database. Constancy is found to remain preserved in this case. But for space constraint we cannot provide this result.

Algorithm 1, based on Coiflet and Daubechies wavelet packet features, and the Algorithm 2, based on directional features through Radon transform, both have been tested using different number of training images. Classification was conducted in the LDA space using k-NN classifier based on $L1$ norm measure and Mahalanobis distance measure. The results are shown in Table 1, Table 2 and Table 3.It is seen that the coiflet based Algorithm 1 and the Radon transform based Algorithm 2 have nearly similar performance. Coiflet based classification is found to be superior to Daubechies wavelet based classification for all the three data bases. The justification is coiflet has zero wavelet and zero scaling function moments in addition to one zero moment for orthogonality.

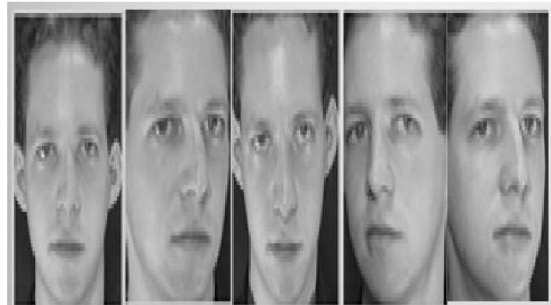

Fig.5 Sample images of ORL Database

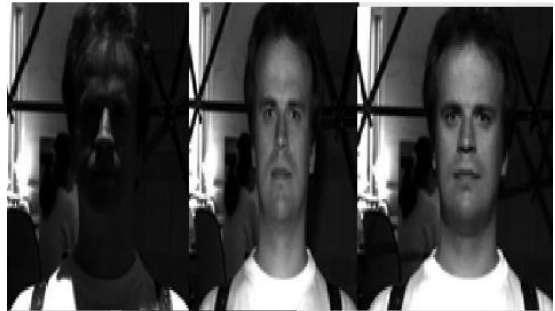

Fig.6 Sample images of Yale Database

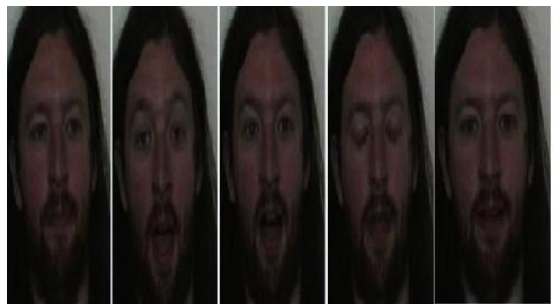

Fig.7 Sample images of Essex Grimace Database

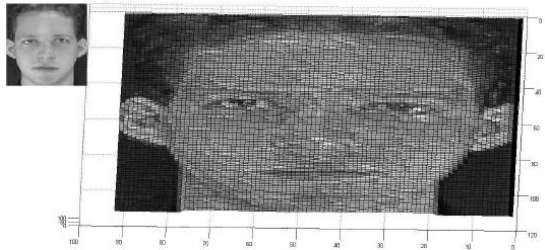

Fig.8 Image and its depth map

Table 2: Daubechies Wavelet Based Classification

| Sl.No. | Data-base | No.of Tr.Images | Classi-fication | Recog-nition % |
|---|---|---|---|---|
| 1 | ORL | 5 | L1 norm | 90.0 |
| 2 | ORL | 5 | Mahalanobis | 90.0 |
| 3 | ORL | 4 | L1 norm | 93.3 |
| 4 | ORL | 4 | Mahalanobis | 95.0 |
| 5 | ORL | 3 | L1 norm | 92.8 |
| 6 | ORL | 3 | Mahalanobis | 88.5 |
| 7 | ORL | 2 | L1 norm | 86.25 |
| 8 | ORL | 2 | Mahalanobis | 81.25 |
| 9 | Yale | 2 | L1 norm | 100.0 |
| 10 | Yale | 2 | Mahalanobis | 100.0 |
| 11 | Essex-Grimace | 2 | L1 norm | 100.0 |
| 12 | Essex-Grimace | 2 | Mahalanobis | 100.0 |
| 13 | Essex-Grimace | 1 | L1 norm | 100.0 |
| 14 | Essex-Grimace | 1 | Mahalanobis | 73.3 |

Table 3: Radon Transform Based Classification (Algorithm 2)

| Sl.No. | Data-base | No.of Tr.Images | Classi-fication | Recog-nition % |
|---|---|---|---|---|
| 1 | ORL | 5 | L1 norm | 98.0 |
| 2 | ORL | 5 | Mahalanobis | 100.0 |
| 3 | ORL | 4 | L1 norm | 91.6 |
| 4 | ORL | 4 | Mahalanobis | 96.3 |
| 5 | ORL | 3 | L1 norm | 91.4 |
| 6 | ORL | 3 | Mahalanobis | 94.2 |
| 7 | ORL | 2 | L1 norm | 81.2 |
| 8 | ORL | 2 | Mahalanobis | 86.2 |
| 9 | Yale | 2 | L1 norm | 100.0 |
| 10 | Yale | 2 | Mahalanobis | 100.0 |
| 11 | Essex-Grimace | 2 | L1 norm | 100.0 |
| 12 | Essex-Grimace | 2 | Mahalanobis | 100.0 |

This provides the superior results compared to wavelets with only zero wavelet moments. Comparison in Table 4 shows the proposed method (Algorithm 1, Algorithm 2) is better than the Curvelet based PCA [1] and DWT based [2] method, even at lower number of training images.

## 6. Conclusion

We have proposed two different algorithms with two different kinds of features and these features are reasonably robust against illumination variation and orientation. These facts are supported by experimental results based on images of Yale and ORL data bases respectively. The choice of Yale and ORL data bases are due to large variation in illumination and orientation. Tests on Essex-Grimace database proves that both the proposed algorithms are robust against variations in expression and provides 100% Recognition Rate for just 2 training images per class. Tests on the Yale database proves that the proposed algorithms, particularly the second algorithm based on Radon Transform, is robust against major variations in illumination for just 2 training images per class and provides 100% Recognition Rate. ORL database whose images have minor variations of pose, expression and illumination has also been tested and a recognition rate of 100% has been achieved using the first algorithm and 98% recognition rate has been achieved using the second algorithm.

Table 4: Comparison of Results

| Sl.No. | Data-base | No.of Tr.Images | Classi-fication | Recog-nition % |
|---|---|---|---|---|
| 1 | ORL | 5 | Curvelet Based PCA[1] | 96.6 |
| 2 | Essex-Grimace | 8 | Curvelet Based PCA[1] | 100.0 |
| 3 | ORL | 5 | DWT & Image Comparison[2] | 99.5 |
| 4 | ORL | 4 | Proposed Algo-1 | 100.0 |
| 5 | Yale | 2 | Proposed Algo-1 | 100.0 |
| 6 | Essex-Grimace | 2 | Proposed Algo-1 | 100.0 |
| 7 | ORL | 5 | Proposed Algo-2 | 98.0 |
| 8 | Yale | 2 | Proposed Algo-2 | 100.0 |
| 9 | Essex-Grimace | 2 | Proposed Algo-2 | 100.0 |

**Sambhunath Biswas** obtained the M.Sc. degree in Physics, and Ph.D. in Radiophysics and Electronics, from the University of Calcutta in 1973 and 2001 respectively. In 1983-85 he completed courses of M. Tech. (Computer Science) from Indian Statistical Institute, Calcutta.His doctoral thesis is on *Image Data Compression*. He was a UNDP Fellow at MIT, USA to study machine vision in 1988-89 and was exposed to research in the Artificial Intelligence Laboratory. He visited the Australian National University at Canberra in 1995 and joined the school on wavelets theory and its aplications. In 2001, he became a representative of the Government of India to visit China for holding talks about collaborative research projects in different Chinese Universities.He started his career in Electrical Industries, in the beginning as a graduate engineering treinee and, then as a design and development engineer. He, at present, is a system analyst, GR-I (in the rank of Professor) at the Machine Intelligence Unit in Indian Statistical Institute, Calcutta where he is engaged in research and teaching. He is also an external examiner of the Department of Computer Science at Jadavpur University in Calcutta.He has published several research articles in International Journals of repute and is a member of IUPRAI (Indian Unit of Pattern Recognition and Artificial Intelligence). He is the principal author of a book titled *Bezier and Splines in Image Processing and Machine Vision*, published by Springer, London.

**Amrita Biswas** obtained B.Tech Degree in Electronics and Communication engineering from Sikkim Manipal Institute of Technology in 2004 and M.Tech in Digital Electronics and Advanced Communication from Sikkim Manipal University in 2005.She is presently working in SMIT as an Associate Professor.She is also pursuing her PhD.Her current research areas include Pattern Recognition and Image Processing.